\documentclass{article}

\usepackage[preprint]{neurips_2023}

\usepackage[utf8]{inputenc} 
\usepackage[T1]{fontenc}    
\usepackage{hyperref}       
\usepackage{booktabs}       
\usepackage{amsfonts}       
\usepackage{nicefrac}       
\usepackage{microtype}      
\usepackage{xcolor}         
\usepackage{times}
\usepackage{soul}
\usepackage[small]{caption}
\usepackage{graphicx}
\usepackage{amsmath}
\usepackage{amsthm}
\usepackage{booktabs}
\usepackage{algorithm}
\usepackage{algorithmic}
\usepackage{multirow}
\usepackage{lipsum}    
\usepackage{booktabs}
\usepackage{amssymb}
\usepackage{amsmath}
\usepackage{xcolor}
\usepackage{subcaption}
\usepackage{natbib}

\newcommand{\citea}[1]{(\cite{#1})}

\newcommand{\wm}[1]{{#1}}

\newcommand{\llv}[1]{\textbf{\textcolor{orange}{}}}
\newcommand{\awais}[1]{\textbf{\textcolor{blue}{}}}

\title{FROD: Robust Object Detection for Free}

\author{%
  Muhammad Awais\thanks{This work was carried out at Sony AI. More details at \href{https://awaisrauf.github.io/frod}{awaisrauf.github.io/frod}.} \\
  Sony AI\\
  Japan\\
  \texttt{iawaisrauf@gmail.com} \\
  \And
  Weiming Zhuang \\
  Sony AI \\
  Japan \\
  \texttt{weiming.zhuang@sony.com} \\
  \AND
   Lingjuan Lyu \\
  Sony AI \\
  Japan \\
  \texttt{lingjuan.lv@sony.com } \\
  \And
  Sung-Ho Bae\thanks{Corresponding author} \\
  Kyung-Hee University \\
  South Korea \\
  \texttt{shbae@khu.ac.kr} \\
}

\begin{document}

\maketitle

\begin{abstract}
Object detection is a critical task in computer vision and has become an integral component of numerous critical systems. However, state-of-the-art object detectors, similar to their classification counterparts, are susceptible to small adversarial perturbations that can significantly alter their normal behavior. Unlike classification, the robustness of object detectors has not been thoroughly explored. In this work, we take the initial step towards bridging the gap between the robustness of classification and object detection by leveraging adversarially trained classification models. Merely utilizing adversarially trained models as backbones for object detection does not result in robustness. We propose effective modifications to the classification-based backbone to instill robustness in object detection without incurring any computational overhead. To further enhance the robustness achieved by the proposed modified backbone, we introduce two lightweight components: imitation loss and delayed adversarial training. Extensive experiments on the MS-COCO and Pascal VOC datasets are conducted to demonstrate the effectiveness of our proposed approach.
\end{abstract}

\section{Introduction}
\label{sec:intro}

Deep learning models have demonstrated remarkable performance in various computer vision tasks, including image recognition \citea{he2016deep, krizhevsky2017imagenet, xie2020self}, object detection \citea{girshick2015region, ren2015faster, redmon2016you, kang2022survey}, and semantic segmentation \citea{chen2017deeplab, 9356353}. Despite their achievements, deep learning models are susceptible to adversarial attacks, which involve subtle and imperceptible alterations in the input space \citea{biggio2013evasion, szegedy2013intriguing, carlini2017towards}. These attacks have raised significant concerns regarding security and robustness \citea{brown2017adversarial, ma2021understanding}. While considerable efforts have been made to counteract these attacks, the majority of research has primarily focused on defending image classification models.

Object detection plays a vital role in numerous real-world applications, including autonomous vehicles and tracking systems \citea{zou2019object}. Its objective is to accurately locate and classify multiple objects of various scales within an input image. Extensive research efforts have been dedicated to enhancing object detection models, resulting in notable advancements \citea{ren2015faster, redmon2016you, lin2017focal, tan2020efficientdet, carion2020end}. However, despite their state-of-the-art performance, these models are susceptible to adversarial attacks that can undermine their reliability \citea{xie2017adversarial, song2018physical}. These attacks have the potential to cause erroneous object localization and recognition, leading to significant challenges when deploying such models in real-world scenarios \citea{song2018physical, liu2018dpatch, wu2020making}.

However, in contrast to the plethora of defense methods developed to counter attacks on image classification models \citea{madry2017towards, uesato2019labels, wong2020fast, shafahi2019adversarial}, the research on defending against adversarial attacks in object detection remains relatively limited \citea{zhang2019towards, chen2021class}. These existing works propose modified versions of adversarial training specifically tailored for object detection. For example, \cite{chen2021class} introduced improved calibration techniques for class-wise and object-wise losses in adversarial training. However, these methods require training robust object detectors through computationally intensive adversarial training and acquiring robustness from scratch. Additionally, these approaches do not directly leverage the advancements made in enhancing the robustness of classification models.

In this work, our objective is to leverage pre-trained adversarially robust models from image classification to enhance the robustness of object detection, while maintaining a minimal computational overhead compared to standard training. Specifically, we propose replacing the standard backbone of object detection with a robust backbone obtained from an adversarially trained classification model. However, a straightforward switch of backbones does not yield the desired robustness, as standard training leads to catastrophic forgetting of the backbone's robustness. To address this issue, we introduce Free Robust Object Detection (FROD), a method that incorporates simple modifications based on robust backbones. We refer to our approach as "free" because it instills robustness without incurring any additional computational cost compared to standard training. To further enhance the robustness of our method, we introduce two new components during the training process: imitation loss and delayed adversarial training. This approach, called FROD-DAT, is computationally efficient, as the adversarial training is performed only at the end of the training process using a single-step adversary.

We conducted extensive qualitative and quantitative experiments to evaluate the effectiveness of our method in instilling robustness and improving clean performance. The results demonstrate that our method achieves SOTA-comparable robustness while achieving significantly higher clean mAP. 

\begin{figure}[t]
    \centering
    \includegraphics[width=1\textwidth]{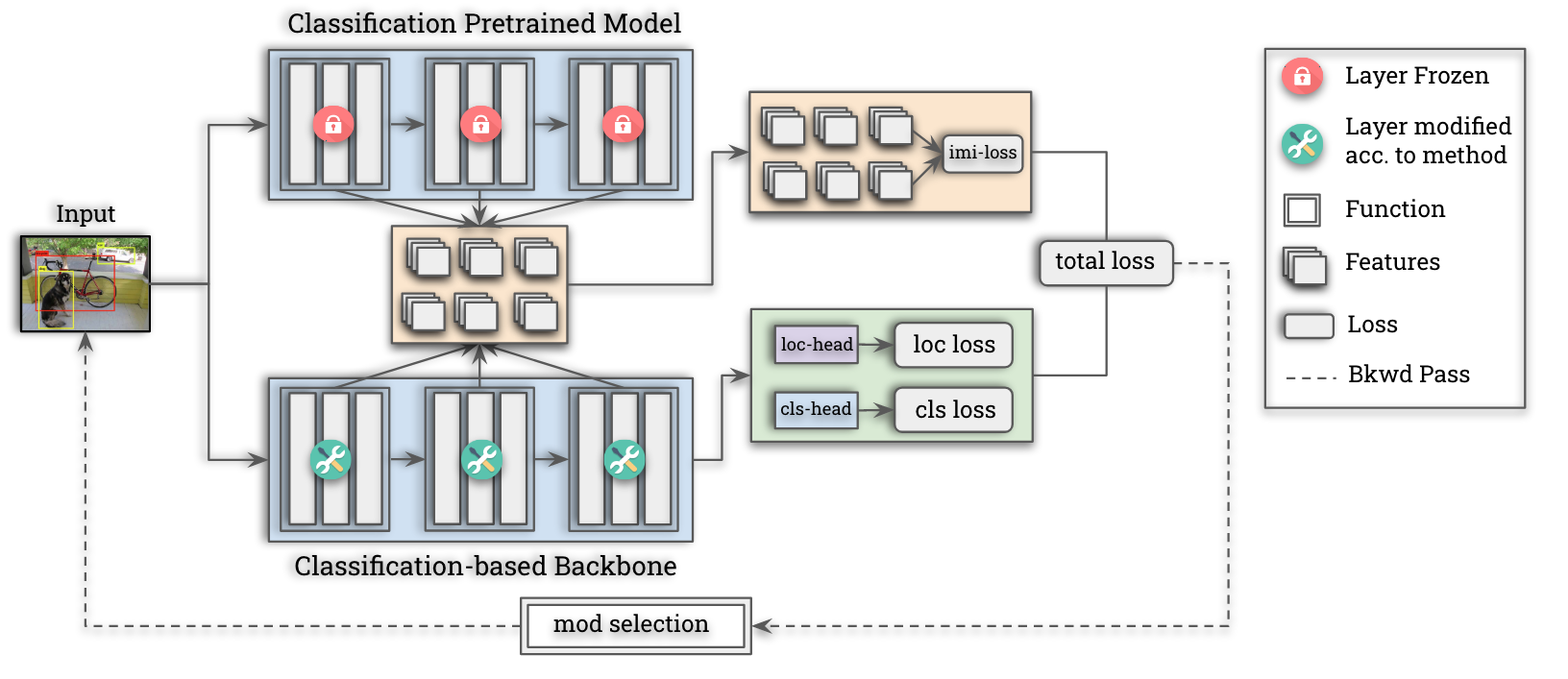}
    \caption{Overview of our proposed method. We leverage classification-based pre-trained models as backbones (blue) and modify them for effective robustness transfer. The fixed robust model provides robust features for imitation during training. The selection module (sel-mod) determines the training modes based on the chosen method (FROD or FROD-DAT).}
    \vspace{-10pt}
    \label{fig:overview}
\end{figure}

\section{Related Work}  

\textbf{Adversarial Attacks and Robustness}. Since the seminal work on adversarial vulnerability of neural networks by \cite{szegedy2013intriguing}, extensive research has focused on constructing adversarial attacks \citea{goodfellow2014explaining,moosavi2016deepfool,papernot2017practical} and developing defenses against them~\citea{madry2017towards,katz2017reluplex,weng2018towards, muhammad2022survey}. However, most of these efforts have primarily targeted general classification models. Notable attack methods include Fast Gradient Sign Method (FGSM) by \cite{goodfellow2014explaining} and Projected Gradient Sign (PGD) by \cite{madry2017towards}, widely used for evaluating white-box robustness. Adversarial training by \cite{madry2017towards}, on the other hand, represents a generic approach that offers effectiveness against various adversarial attacks~\citea{athalye2018obfuscated}.

\noindent \textbf{Adversarial Robustness for Object Detection}. Significant advancements have been made in object detection \citea{ren2015faster,redmon2016you,lin2017focal,tan2020efficientdet,carion2020end}, with notable architectures such as single-stage~\citea{lin2017focal} and two-stage~\citea{ren2015faster} detectors. However, similar to classification models, state-of-the-art object detection models have been found to be vulnerable to adversarial perturbations~\citea {xie2017adversarial,song2018physical,liu2018dpatch,wu2020making}. In fact, slight modifications in physical environments can deceive deployed vision systems~\citea{song2018physical}. Our approach addresses this vulnerability and is applicable to both single-stage and two-stage object detection architectures.

Works focusing on adversarial robustness for object detection are relatively scarce. Some recent papers have introduced modified versions of adversarial training specifically tailored for object detection~\citea{zhang2019towards,chen2021class}. \cite{zhang2019towards} formulated object detection as a multi-task learning problem, emphasizing the misalignment between classification and localization loss gradients. They proposed MTD-A, which incorporated a task-oriented domain constraint based on the maximization of either loss. To expedite the training, they employed a FastAT-like strategy. On the other hand, \cite{chen2021class} identified limitations in MTD, particularly its failure to account for the multi-class and multi-object nature of object detection. They addressed the issue by decomposing the loss based on objects, applying clipping, and normalizing the loss of each class for balanced influence, known as Class-Wise Adversarial Training (CWAT).

In contrast, our approach significantly differs from these prior works, as we build upon the adversarially pre-trained classifiers. Nonetheless, our approach shares some similarities with these approaches, as we also employ adversarial training within our two-phase framework.

\noindent \textbf{Robust Features in Classification.} In the realm of classification, recent studies have highlighted the divergent nature of features learned by adversarially trained models compared to their standard counterparts~\citea{zhu2021towards,ilyas2019adversarial,santurkar2019image,tsipras2018robustness,engstrom2019adversarial, dong2022zood}, thereby demonstrating the potential for feature transferability. Existing approaches for feature transfer include transfer learning-based methods~\citea{shafahi2019adversarially,hendrycks2019pretraining} and distillation-based techniques~\citea{awais2021adversarial,shafahi2019adversarial,zhang2019you,wong2020fast,muhammad2021mixacm}. While our work differentiates itself by focusing on leveraging classification models to enhance object detection robustness, it draws inspiration from these prior works. Our method is particularly useful as countless general classification-based pre-trained models are readily available through numerous open-source projects such as HuggingFace. 

\begin{table}[t]
\resizebox{0.95\textwidth}{!}{%
\centering
\begin{tabular}{lcp{0.05\textwidth}p{0.05\textwidth}p{0.05\textwidth}p{0.05\textwidth}ccc}
\toprule
        \multirow{2}{*}{\textbf{Method}}   & \multirow{2}{*}{\textbf{Clean}}     & \multicolumn{2}{c}{\textbf{FGSM}}  & \multicolumn{2}{c}{\textbf{PGD-10}} & \multicolumn{2}{c}{\textbf{Add. Steps}} & \textbf{Comp.} \\
          \cmidrule(lr){3-4}\cmidrule(lr){5-6} \cmidrule(lr){7-8}&&\textbf{$A_{cls}$}&\textbf{$A_{reg}$}&\textbf{$A_{cls}$}&\textbf{$A_{reg}$} &Fwd & Bkwd& \textbf{Overhead}\\
\midrule
\textbf{STD} & \textbf{0.752} & 0.162 & 0.25 & 0.012 & 0.043 & 0&0&1$\times$ \\
\textbf{MTD-fast}~\citea{zhu2021towards} & 0.466 & 0.311 & 0.418 & 0.221 & 0.351 & 1& 1&2.78$\times$\\
\textbf{TOAT-6} \citea{chen2021class}& 0.430 & 0.300 & 0.397 & 0.218 & 0.334 & 7& 7& 13.47$\times$ \\
\textbf{CWAT}~\cite{chen2021class} & 0.513 & 0.325 & 0.433 & 0.224 & 0.367 & 7 & 7 &13.47$\times$\\
\midrule
\textbf{FROD } (Ours) & {0.671} & {0.498} & {0.581} & 0.202 & 0.358  & 0 & 0 & 1$\times$ \\
\textbf{FROD-DAT} (Ours) & 0.648 & \textbf{0.534}  & \textbf{0.593} & \textbf{0.252} & \textbf{0.419} & 1 & 1 &1.91$\times$  \\
\bottomrule
\end{tabular}
}
\caption{A comparison of adversarial robustness and clean mAP on Pascal VOC for RetinaNet trained with our method and previous approaches. The computational overhead compared to standard training is also shown, indicated by the additional forward (Fwd) and backward (Bkwd) steps required by each method and training time comparison. The computational overhead is estimated empirically.}
\vspace{-20pt}
\label{table:pascal_free}
\end{table}

\section{Problem Formulation}
An object detector maps an input image $x$ to a set of \wm{$K$} objects represented with bounding boxes $b_k$ and a probability vector \wm{$p_k$} for each object spanning $C$ classes: \wm{$f(x)$}$ \to \{p_k, b_k\}_{k=1}^{K}$. After that, \wm{it uses} Non-Maximum Suppression (NMS)~\citea{rosenfeld1971edge} to remove redundant detection boxes.
\wm{An object detector contains a backbone parameterized by $\theta$ and classification and localization heads \footnote{Two-stage detectors contain only a classification head, but single-stage detectors contain an additional localization head. We formulate the problem based on single-stage detectors, but our method works on both single-stage and two-stage detectors.} parameterized by $\omega$. Object detection aims to estimate $\{\theta, \omega\}$ by minimizing a loss: 
\begin{equation}
    \min_{\theta, \omega} \mathcal{L} (f_{\theta, \omega}(x), \{y_k, b_k\}),
\end{equation}
where $y_k$ is the class label for object $k$ and $\mathcal{L}(\cdot)$ is the loss function. The backbone is for extracting features, and its parameters $\theta$ are commonly initialized 
from a model pre-trained on large-scale classification datasets like ImageNet \citea{deng2009imagenet}. The loss function $\mathcal{L}(\cdot)$ is a combination of a classification loss $\mathcal{L}_{cls}$ and a localization loss $\mathcal{L}_{loc}$, we then further formulate the objective as follows:
\begin{equation}
    \min_{\theta, \omega} \mathcal{L}_{cls} (f_{\theta, \omega}(x), y_k) + \mathcal{L}_{loc} (f_{\theta, \omega}(x), b_k).
\end{equation}
}

We further consider the robustness of object detectors. The robustness of an object detector is measured by the performance of the model on a perturbed test set. For adversarial robustness, the perturbation $\delta$ is found by iteratively solving the following objective: $\delta = \max_{\|\delta\|_p \leq \epsilon} \mathcal{L}(x,{y, b}) $,
where $\epsilon$ is the perturbation budget and $\mathcal{L}$ can be classification loss, localization loss, or a combination of both. Fast Gradient Sign Method (FGSM) ~\citea{goodfellow2014explaining} approximated it for $\ell_{\infty}$ and has the following closed form: $\delta_{FGSM} = \epsilon \cdot \text{sign}(\mathcal{L})$. A stronger and standard evaluation attack called Projected Gradient Descent (PGD) attack~\citea{madry2017towards} is based on an iterative solution of this objective.

\section{Methodology}
\label{sec:Methodology}

In this section, we begin by discussing effective strategies for utilizing the robust model in the object detection framework. Next, we introduce an imitation loss mechanism designed to preserve the robustness of the object detector when using a fixed robust backbone. Finally, we present a lightweight technique to enhance the overall robustness of the object detector through the incorporation of delayed adversarial training.

 \subsection{The Case of Catastrophic Forgetting of Robustness}

Most modern object detectors employ features extracted from a backbone pre-trained on classification datasets. These features are then fed to small classification and regression networks. The backbone plays a crucial role as both classification and localization networks share the same backbone and use its extracted features. Existing object detection methods mostly initialize the backbone with a pre-trained model (on a dataset like ImageNet \cite{deng2009imagenet}) and train the model on object detection datasets. We hypothesize that adopting a pre-trained \textit{robust} backbone could enhance the robustness of object detection models. 

To test this hypothesis, a basic question is: can we achieve robustness by simply switching a normally trained backbone with a robust counterpart? To answer this question, we perform an experiment with RetinaNet. We switch the standard backbone with a pre-trained robust counterpart while keeping all other settings intact, three blocks being retrained and frozen BatchNorm~\footnote{Referecne code: https://github.com/pytorch/vision/tree/main/ references/detection}. The results are shown in Table~\ref{tab:backbone_exps}(a). The results demonstrate an interesting case of catastrophic forgetting of robustness. 

\subsection{Effective Utilization of Robust Backbone}
\label{sec:initialization} 
To efficiently leverage robust pre-trained backbones, we introduce two light modifications: retraining fewer layers and updating the batch normalization layers (BatchNorm) of backbones on the new dataset. It is important to note these two modifications do not increase the training or inference time of a model. The details of these two modifications are as follows. 

\begin{table}[t]
\centering
\resizebox{\textwidth}{!}{%
\begin{tabular}{lcccccccccc}
\toprule
\multicolumn{1}{l}{\multirow{3}{*}{\textbf{attack}}} &  \multicolumn{2}{c}{\textbf{Clean}} &  \multicolumn{4}{c}{\textbf{FGSM}} &\multicolumn{4}{c}{\textbf{PGD-10}} \\
\cmidrule(lr){2-3}\cmidrule(lr){4-7}\cmidrule(lr){8-11}
 &\multirow{2}{*}{{0.5}} & \multirow{2}{*}{{0.5:0.95}}  & \multicolumn{2}{c}{$A_{cls}$}  &\multicolumn{2}{c}{$A_{reg}$} &\multicolumn{2}{c}{$A_{cls}$}&\multicolumn{2}{c}{$A_{reg}$} \\
 \cmidrule(lr){4-5}\cmidrule(lr){6-7}\cmidrule(lr){8-9}\cmidrule(lr){10-11}
 &  &   &0.5  &0.5:0.95 &0.5&0.5:0.95 &0.5 &0.5:0.95&0.5&0.5:0.95 \\
\toprule
\textbf{STD} & - & \textbf{0.451} & 0.133 & - & 0.167 & - & 0.030 & - &  0.029 & - \\

\textbf{MTD}{\textsuperscript{1}} & & 0.190  & 0.127 & - &  0.146 & - & 0.110 & -  & 0.135 &- \\
\textbf{MTD-fast} & - & 0.242 & 0.167 & - & 0.182 & - & 0.130 & - & 0.134 & -\\
\textbf{TOAT-6} & - & 0.182 & 0.120 & - &  0.148 & - &  0.098 & - & 0.123 & - \\

\textbf{CWAT} & - & 0.237 & 0.168 & - & 0.189 & - & 0.142 & - & 0.155 & -- \\
\midrule
\textbf{FROD}    & 0.415 & \underline{0.249}   & \textbf{0.292} & 0.169 & \textbf{0.352}& 0.184   & 0.121  & 0.068 & \underline{0.246} & 0.100 \\

\textbf{FROD-DAT}  & 0.356 & 0.216  & \underline{0.275} & 0.163 & \underline{0.318} & 0.173  & \textbf{0.153} & 0.088  & \textbf{0.253} & 0.110 \\
\bottomrule
\end{tabular}
}
\caption{A comparison of adversarial robustness and clean mAP on MS-COCO for object detectors trained with various robustness algorithms. The adversarial training is performed with a perturbation budget of $\epsilon=8$.}
\label{table:cocoforfree}
\vspace{-10pt}
\end{table}

\begin{figure}[htbp]
  \begin{subfigure}[b]{0.45\linewidth}
    \centering
    \includegraphics[width=\textwidth]{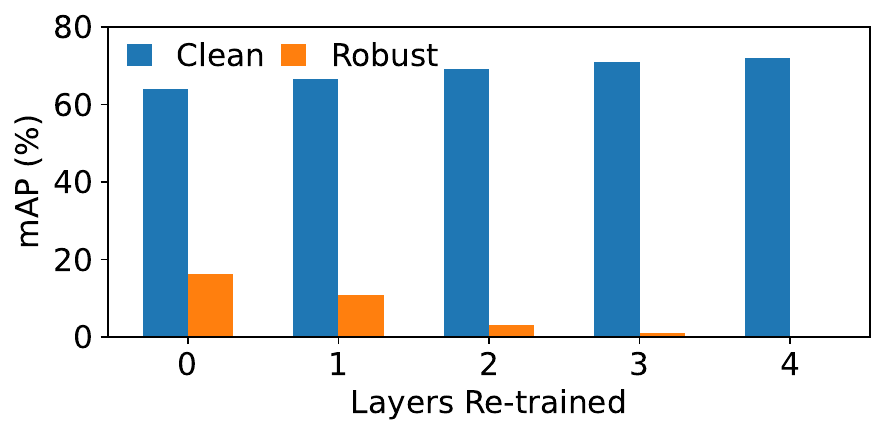}
  \end{subfigure}%
  \begin{subfigure}[b]{0.45\linewidth}
    \centering
    \includegraphics[width=\textwidth]{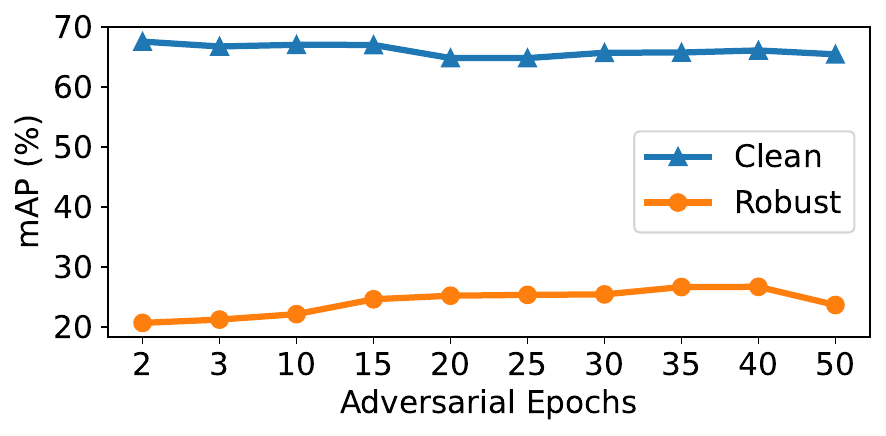}
  \end{subfigure}
  \caption{
  (a) A comparison of robustness and clean mAP as the number of retrained blocks increases. Robustness decreases as the number of retrained blocks in the backbone increases. (b) The impact of starting adversarial training at different epochs. }
  \vspace{-10pt}
  \label{backbone_exps_plots}
\end{figure}

\textbf{Retraining of Layers. }
This problem has been studied for classification~\citea{li2017learning,shafahi2019adversarially} and previous results show that the backbone forgets its robustness when retrained on standard examples. The number of backbone layers retrained is vital for this problem~\citea{shafahi2019adversarially}. To understand the role of layers in the preservation of robustness, we divided the backbone layers into four blocks, following the original ResNet~\citea{krizhevsky2017imagenet} configuration. 
Then, we performed an experiment where we progressively increased the number of blocks in the backbone that are being retrained. The details of settings are in Section~\ref{sec:experiment_settings}. We start with no block retrained (0) to all the blocks retrained (4). These experiments are performed with a RetinaNet trained on the Pascal VOC dataset. The results are shown in Figure~\ref{backbone_exps_plots}(a). The results suggest a clear trend for both robustness and clean mAP: the more blocks retrained, the less robustness, and the higher the clean accuracy. 
 
 Based on our empirical study, we concluded that retraining zero or one block is the optimal configuration to preserve robustness. We also observed that retraining of blocks also acts as a trade-off between robustness and clean mAP. Specifically, the more layers retrained, the lesser the robustness and the higher the accuracy. 
 
\textbf{Updating BatchNorm. }
Batch Normalization (BatchNorm) has shown to have a significant role in the adversarial robustness~\citea{benz2020batch,xie2020adversarial, muhammad2023adversarial, awais2020towards, awais2020revisiting}. BatchNorm keeps track of batch statistics during training to estimate the population statistics. These estimates are employed during inference. It has been shown that these statistics play a crucial role in the overall robustness of a model. To understand the role of batch statistics of the backbone in object detection, we perform an experiment with frozen and non-frozen BatchNorm layers. For this experiment, we selected the default settings stated in Section~\ref{sec:experiment_settings}.

Table~\ref{tab:backbone_exps}(b) shows that updating BatchNorm improves the robustness of object detection significantly. Based on these results, we propose to update batch statistics since they are data-dependent. This is a crucial insight for the preservation of robustness as most object detection methods freeze BatchNorm in the backbone\citea{wu2019detectron2}. 

\begin{table}[htbp]
  \begin{subtable}[b]{0.5\linewidth}
    \centering
    \begin{tabular}{ccc}
    \toprule
        \textbf{Backbone} & \textbf{Clean mAP} & \textbf{Robust mAP}  \\
    \toprule
       Normal  & 0.648 & 0.0001\\
       Robust &  0.596 & 0.0059\\
    \bottomrule
    \end{tabular}
    \caption{}
  \end{subtable}%
  \begin{subtable}[b]{0.5\linewidth}
    \centering
    \begin{tabular}{ccccc}
        \toprule
         \textbf{Update BN?} & \textbf{Normal mAP} & \textbf{Robust mAP}\\
        \toprule
         Yes  & 0.671    & 0.203 \\
         No & 0.641    & 0.102\\
         \bottomrule
    \end{tabular}
    \caption{}
  \end{subtable}
  \caption{ (a) Does naively switching a standard backbone with a robust backbone yield any robustness? The table compares the normal mAP and PGD10-cls mAP robustness of a model trained with a standard and robust backbone. (b) The impact of updating the BatchNorm layer on the robustness of an object detector.}
  \label{tab:backbone_exps}
\end{table}

\subsection{Imitation of Robust Features}

The preceding section demonstrates that replacing a standard backbone with a robust one along with our proposed approach can enhance the overall robustness of object detectors. However, preserving robustness entirely by freezing certain backbone layers poses a challenge, as it restricts the ability of the backbone (trained on classification) to adopt new concepts and knowledge from the new object detection dataset. To address this issue, we propose a new approach using an imitation loss, as illustrated in Figure~\ref{fig:overview}. Our approach allows more flexibility in the backbone. By maintaining a frozen copy of the pre-trained backbone, we can leverage its robust features to regularize the updates of the backbone model. The formulation of the imitation loss  is defined as follows,
$
\mathcal{L}_{imi} = \sum_{l \in L}\|f_{\theta}^l(x) - f_{\theta'}^l(x)\|_p,     
$
where $l$ is a block in the backbone, $f_{\theta}$ is the backbone, $f_{\theta'}$ is the fixed backbone, and $p$ is the norm used.

\subsection{Efficient Delayed Adversarial Training}
The previous sections present approaches for the utilization of the pre-trained backbone for free robustness. However, free robustness is limited as robust backbones are pre-trained on a different dataset for a different task to solve an entirely different problem. In this section, we further study how to utilize a pre-trained robust backbone more effectively with adversarial training while maintaining the efficiency of our method.

To this end, we propose a two-phase approach consisting of regular training and a single-step, delayed adversarial training. Our method first trains object detectors on normal examples for $t_1$ epochs. Second, the training is switched to the single-step-based adversarial examples~\citea{wong2020fast} for $t_2$ epochs. 
This mechanism results in a more robust model at the cost of significantly less computation compared with other adversarial training methods. The objective for the standard training phase is as follows, 

\begin{equation}
     \arg \min_{\theta, \omega} \bigg[ \mathcal{L}_{cls}(x, y; \theta, \omega) + \mathcal{L}_{loc}(x, b; \theta, \omega)  + \lambda \cdot \mathcal{L}_{imi}(x, \theta, \theta') \bigg].
    \label{eq:overall_loss_p1}
\end{equation}

Similarly, the objective function for the adversarial training phase is as follows, 
\begin{equation}
     \arg  \min_{\theta, \omega} \bigg[\max_{\|\delta\|_p \leq \epsilon}  \mathcal{L}_{cls}(x+\delta, y; \theta, \omega) +  \mathcal{L}_{loc}(x+\delta, b; \theta, \omega)  + \lambda \cdot \mathcal{L}_{imi}(x+\delta, \theta, \theta') \bigg],
    \label{eq:overall_loss_p2}
\end{equation}

where $\delta$ is adversarial perturbation found by maximizing the classification loss and $\epsilon$ is the perturbation budget. We utilized the single-step method proposed by \cite{wong2020fast}.

\section{Experiments}
\subsection{Experimental Settings}
\label{sec:experiment_settings}

\begin{figure}
    \centering
    \includegraphics[width=1\linewidth]{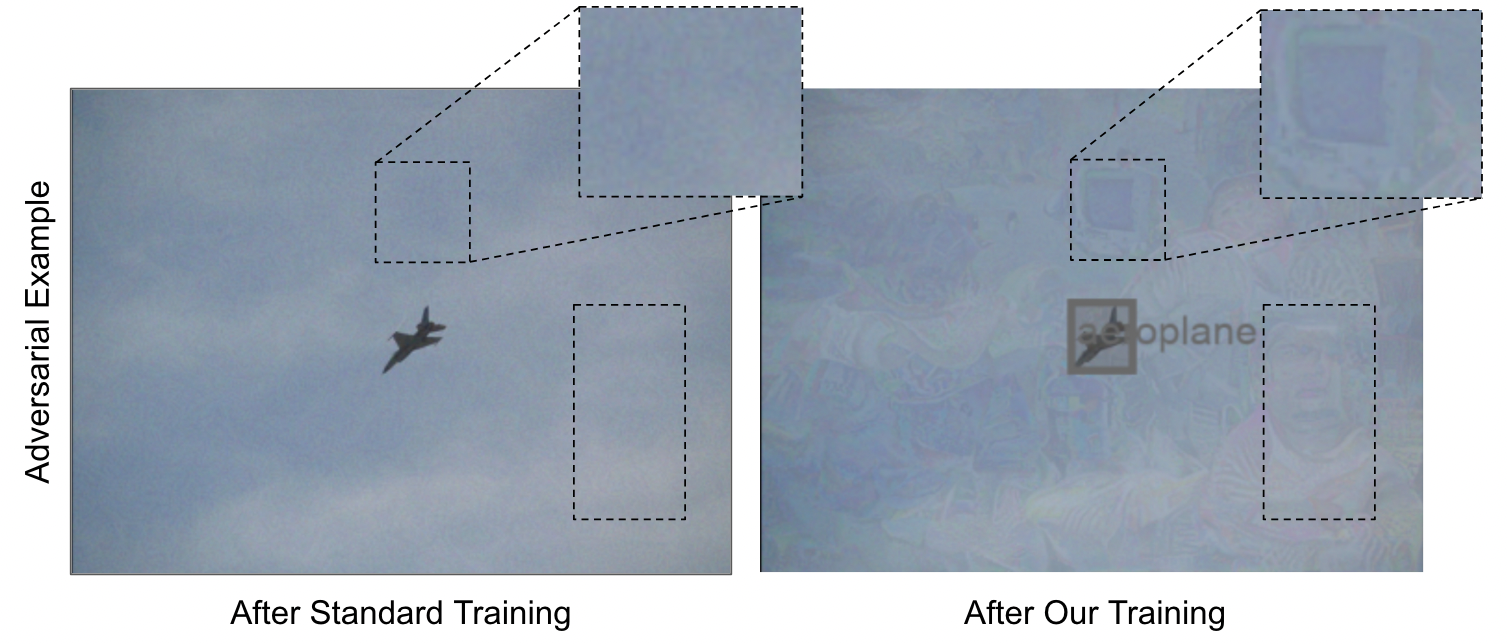}
    \caption{Emergence of human-aligned patterns. Comparison of adversarial distortions crafted for a normal model versus our proposed approach. The distortions generated for the normally trained model are barely perceptible. In contrast, the adversarial attack applied to our method reveals the emergence of human-aligned patterns.}
    \vspace{-10pt}
    \label{fig:hallucinations}
\end{figure}

We evaluate our approach using two widely used object detection datasets: Pascal VOC \citea{everingham2010pascal} and MS-COCO \citea{lin2014microsoft}. For Pascal VOC, we adopt the standard "07+12" protocol for training. This protocol involves using approximately 16,000 images from the combined trainval sets of the 2007 and 2012 datasets, covering 20 different object classes. The test set consists of 4,952 images from the 2007 dataset. As for the MS-COCO dataset, we utilize the train+valminusminival 2014 dataset, which contains around 120,000 images spanning 80 diverse object classes. For the test set, we employ the minival2014 subset, comprising approximately 5,000 images. Our evaluation metric for Pascal VOC is the mean average precision (mAP) with an Intersection-over-Union (IoU) threshold of 0.5. For MS-COCO, we report the mAP at IoU thresholds ranging from 0.5 to 0.95, as per the convention established by \cite{lin2014microsoft}. To represent single-stage detectors, we employ RetinaNet \citea{lin2017focal} with a ResNet50 backbone, while for two-stage detectors, we utilize Faster R-CNN \citea{ren2015faster} with a ResNet50 backbone.
 
All models in our experiments are trained using Stochastic Gradient Descent (SGD) with a momentum of 0.9, weight decay of 1e-4, and a batch size of 4. The initial learning rate is set to 0.04, and it is adjusted based on the batch size and the number of GPUs following the approach by \cite{goyal2017accurate}. For the Pascal VOC dataset, the models are trained for 50 epochs, while for the MS-COCO dataset, the training is performed for 26 epochs. We utilize the training script provided by Torchvision, keeping the original settings unchanged.

To evaluate the robustness of the models, we conduct experiments with an adversarial perturbation budget of $\epsilon=8/255$. We employ a PGD10-based attack constructed based on the classification loss for general robustness evaluation. Additionally, we report the robustness based on classification and regression losses. The following methods are included for comparison.
\textbf{STD}: Object detector trained with standard training on clean images.
\textbf{MTD}: Object detector trained using the robustness algorithm proposed by \cite{zhang2019towards}.
\textbf{TOAT}: Object detector trained using the robustness algorithm proposed by \cite{zhang2019towards}.
\textbf{CWAT}: Object detector trained using the robustness algorithm proposed by \cite{chen2021class}.
\textbf{FROD}: Free version of our proposed method.
\textbf{FROD-DAT}: Our method with imitation loss and delayed adversarial training.

\subsection{Main Results}
\textbf{Comparison with State-of-the-Art. }
To demonstrate the effectiveness of our method, we conducted a comparison with two previous State-of-the-Art (SOTA) robust object detection methods on the Pascal VOC and MS COCO datasets. We first present the results for the Pascal VOC dataset. Table~\ref{table:pascal_free} provides a comprehensive comparison between our method and the previous SOTA methods. It is evident from the table that our Free method (FROD) achieves comparable robustness and clean performance to the SOTA methods, without incurring any additional computational cost. Specifically, our method achieves a robustness of 0.2093, outperforming CWAT (0.224) and TOAT (0.218). Furthermore, our method demonstrates a significantly higher clean mAP of 0.6710, surpassing the values of 0.513 and 0.430 obtained by the previous methods.

Moreover, our FROD-DAT method delivers even more impressive results in terms of clean and robust mAP. Notably, FROD-DAT achieves a clean mAP of 0.648, surpassing the previous SOTA value of 0.513, while simultaneously achieving a robust mAP of 0.2517, outperforming the previous SOTA value of 0.224. This significant improvement in performance highlights the efficacy of our FROD-DAT method.

We further evaluate and compare our method on the challenging MS-COCO dataset~\citep{lin2014microsoft}, which provides a more realistic representation of real-world scenarios. The results of our method are presented in Table~\ref{table:cocoforfree}. Unlike previous works, we report the results of our approach at IoU thresholds of 0.5 and 0.5:0.95 to provide a comprehensive evaluation.

Similar to the observations on the Pascal VOC dataset, our Free method (FROD) demonstrates robustness that is comparable to the previous SOTA methods on MS-COCO. For instance, FROD achieves a robustness of 12.2, outperforming CWAT (14.2) and MTD (13.0). Furthermore, FROD exhibits better clean performance, with a clean mAP of 0.249 compared to 0.237 and 0.190 for CWAT and MTD, respectively. This demonstrates that our method achieves comparable performance to the SOTA methods without any additional computational cost.

Additionally, our FROD-DAT method achieves SOTA-level robustness, with a robust mAP of 0.153 compared to 0.142 and 0.130 for CWAT and MTD, respectively. Remarkably, our FROD-DAT method maintains computational efficiency compared to adversarial training, making it an attractive choice for robust object detection tasks.

\textbf{Computational Complexity.}
We evaluate the computational effectiveness of our method by comparing its complexity and time-per-epoch with previous approaches, as shown in Table~\ref{table:pascal_free}. First, our FROD method does not introduce any additional steps compared to standard training (STD). This ensures that the computational overhead is minimal, allowing for efficient training without compromising performance.
Second, FROD-DAT, our improved method, incorporates adversarial training for less than half of the total training epochs. By reducing the frequency of adversarial training, we strike a balance between robustness and computational efficiency.

Therefore, our method achieves competitive performance while maintaining computational effectiveness, making it a practical and efficient choice for robust object detection tasks.

\begin{figure}
    \centering
    \includegraphics[width=1\textwidth]{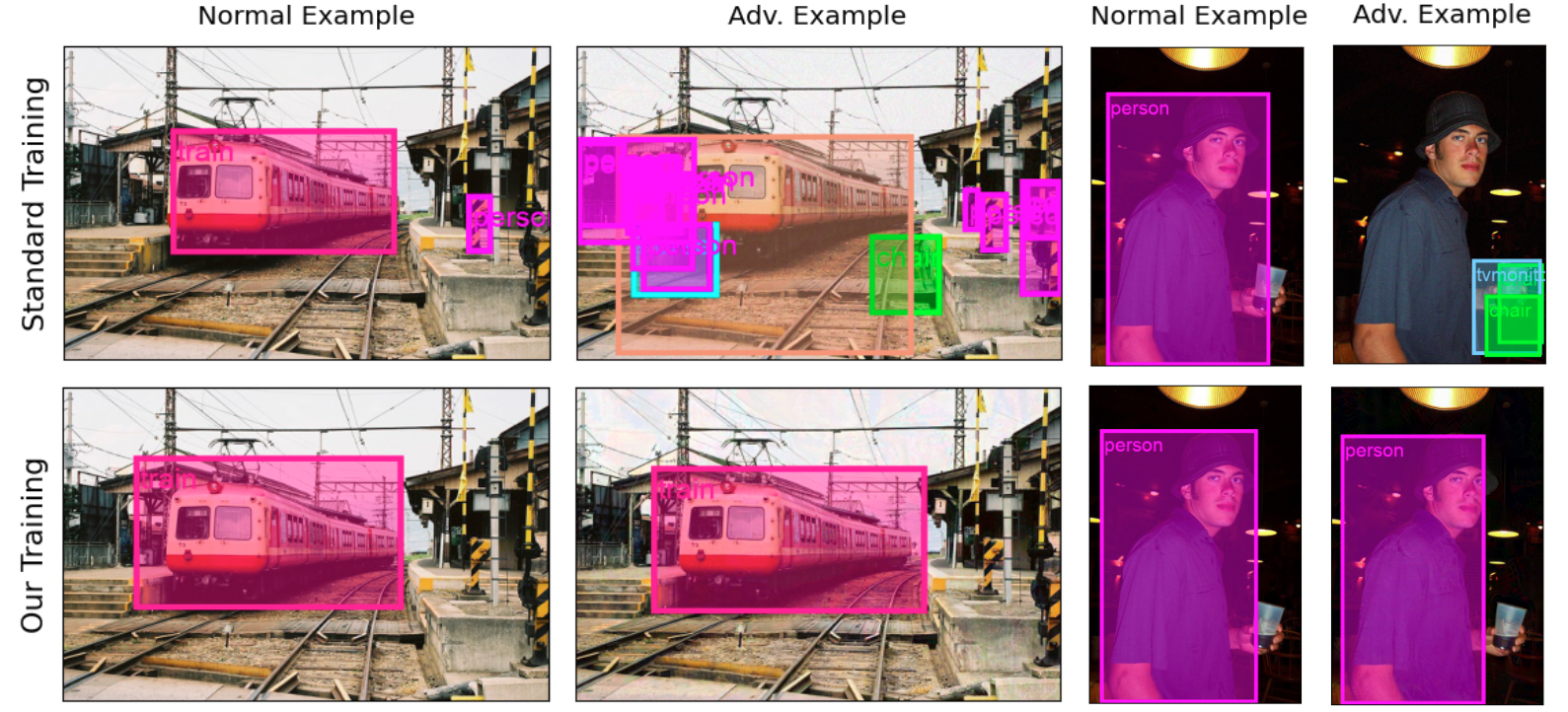}
    \caption{Visual comparison of our method with standard training against PGD-based adversarial attack. The adversarial attack on the normal model leads to hallucinations of non-existent objects. In contrast, our method demonstrates enhanced robustness by mitigating this hallucination effect.}
    \vspace{-5pt}
    \vspace{-10pt}\label{fig:visual_comparison}
    
\end{figure}

\textbf{Single Stage vs Two Stage Detectors. }
Our method is versatile and can be applied to any object detector that utilizes a pre-trained backbone. To demonstrate this, we conducted experiments using both a single-stage detector (RetinaNet) and a two-stage detector (Faster-RCNN) on the Pascal VOC dataset. The results are summarized in Table~\ref{tab:abalation}(b). The table clearly illustrates that our method effectively enhances the robustness of both single-stage and two-stage detectors.

\textbf{Defense Against Transferred Attacks. }
We further test the effectiveness of our method against transferred attacks. Transferred attacks are a type of black-box attacks that are constructed by utilizing another model as a proxy~\cite {goodfellow2014explaining,liu2016delving}. For this purpose, we constructed attacks on Faster RCNN and tested them against RetinaNet. The results are shown in Table~\ref{tab:abalations2}(b). Our method is effective against these attacks.

\subsection{Qualitative Results: Visualizing Adversarial Hallucinations}

\textbf{Emergence of Human-Aligned Patterns. }
Previous research has demonstrated the presence of human-aligned and geometric patterns in adversarial examples crafted for robust models \citea{shafahi2019adversarial, akhtar2021attack}. We observe similar intriguing behavior for our proposed method. In Figure~\ref{fig:hallucinations}, we provide a visual comparison of attacked images between a model trained using a standard method and our proposed approach. The adversarial perturbation crafted for a normally trained image is scarcely perceptible, aligning with the objective of being hidden. However, when targeting an image trained with our method, the resulting perturbations exhibit visible patterns reminiscent of objects found in the training data, such as televisions and humans. Despite these discernible patterns, our method successfully defends against such attacks, showcasing its robustness and effectiveness.

\textbf{Visual Comparison. }
In order to gain deeper insights into the detection results, we conduct a visual comparison between the outcomes of our proposed method and standard training, as depicted in Figure~\ref{fig:visual_comparison}. As illustrated in the figure, the standard model performs well for normal inputs. However, when confronted with adversarial attacks, the standard model exhibits a susceptibility to hallucinating non-existent objects. In stark contrast, our method effectively rectifies these hallucinations and restores the model's sanity, thereby showcasing its ability to defend against adversarial attacks.

\subsection{Ablation Studies}
\begin{table}[htbp]
  \begin{subtable}[b]{0.5\linewidth}
    \centering
    \begin{tabular}{lcc}
    \toprule
             \textbf{Loss}&  \textbf{Clean mAP}  & \textbf{Robust mAP}   \\
                \toprule
     Cls      & 64.80      & 25.17                 \\
     Reg      & 66.05      & 21.63                 \\
     Reg+Cls  & 65.33      & 21.19                 \\
    \bottomrule
    \end{tabular}
    \caption{}
  \end{subtable}%
  \begin{subtable}[b]{0.5\linewidth}
    \centering
    \begin{tabular}{lcc}
     \toprule
            \textbf{Attack}  & \textbf{PGD} & \textbf{PGD} \\
            \textbf{Type}  & \textbf{Cls} & \textbf{Reg} \\
        \toprule
         Direct &   0.2517 & 0.4193 \\ 
       Transferred &  0.5764 & 0.6092\\
       \bottomrule
    \end{tabular}
    \caption{}
  \end{subtable}
  \caption{The impact of updating the BatchNorm layer on the robustness of the object detector. (b) The effectiveness of our method against transferred attacks.}
  \label{tab:abalations2}
\end{table}

\textbf{Sensitivity of Imitation Hyper-parameter. }
In this section, we empirically investigate the role of our proposed imitation loss for the preservation of robustness while allowing flexibility in the backbone.
An essential factor of our imitation loss is the hyperparameter $\lambda$ as defined in Equation~\ref{eq:overall_loss_p1}. This hyperparameter controls the weight of the imitation term in the overall loss. To understand its sensitivity, we experiment with its different values. The results are shown in Table~\ref{tab:abalation}(a). The table shows the relative stability of our method across a range of hyperparameter values.  

\begin{table}[htbp]
  \begin{subtable}[b]{0.5\linewidth}
    \centering
    \begin{tabular}{llccccc}
    \toprule
         $\lambda$     & & \textbf{0.1} & \textbf{0.5} & \textbf{1} \\
    \toprule
    
    w/o Adv. & Clean &   0.653 & 0.656 & 0.652 \\ 
    training & Rob. &   0.093 & 0.099 & 0.095 \\
    \midrule
      w/ Adv.  & Clean & 0.616 & 0.615 & 0.613 \\
        training & Rob. & 0.224 & 0.235 & 0.231 \\

     \bottomrule
    \end{tabular}
    \caption{ }
  \end{subtable}%
  \begin{subtable}[b]{0.5\linewidth}
    \centering
    \begin{tabular}{llcc}
    \toprule
     \textbf{Method} & \textbf{Model} & \textbf{Clean} & \textbf{Robust}  \\
    \toprule
    FROD & RetinaNet  & 0.6710 & 0.2015\\
    & FasterRCNN & 0.6959 & 0.1686\\
    \midrule
    FROD-DAT & RetinaNet   & 0.6480 & 0.2517 \\
    & FasterRCNN  & 0.6693 &  0.2495 \\
    \bottomrule
    \end{tabular}
    \caption{}
  \end{subtable}
  \caption{(a) The impact of increasing imitation loss hyperparameter $\lambda$ on the robustness and clean mAP. (b) Evaluation of our methods for single-stage and two-stage detectors.}
  \label{tab:abalation}
\end{table}

\textbf{Switching to Adversarial Training. } To set the switch points for adversarial training, we performed an experiment where we switched adversarial training at several different points in FROD-based training. As shown in Figure~\ref{backbone_exps_plots}(b), robustness first steadily improves when 
adversarial training starts at different
epochs. However, after a particular position, the improvements saturate and dwindle. For instance, robustness improves starting from epoch 2 to epoch 35. It starts decreasing beyond that point. However, after epoch 30, the improvement in robustness is relatively marginal. Therefore, we select epoch 30 as the starting point for adversarial training. 

\textbf{Misalignment of Tasks in Object Detection. }
Adversarial training crafts adversarial examples using backpropagation with respect to a loss function. Since object detection has a multi-task learning objective function, we need to understand the effect of different loss terms. The previous work~\cite{zhang2019towards} has shown misalignment of the gradient to craft adversarial examples. Hence, we performed experiments with different loss terms to craft FGSM perturbations to understand the role of different loss functions. The results of these experiments are shown in Table~\ref{tab:abalations2}(a). The results show that FGSM perturbation with only classification loss terms is better than regression or both classification and regression loss terms. This case could be because the backbone is already trained with adversarial examples only crafted with classification loss. 

\section{Conclusion}

In this work, we have presented an approach that leverages robust pre-trained classification models to instill adversarial robustness in object detection models. We have found that simply utilizing a classification-based robust pre-trained backbone does not inherently confer robustness in object detection. To overcome this limitation, we have proposed effective modifications in these backbones to harness their robustness for object detection, resulting in robust object detection without incurring extra overhead. To further improve the robustness, we have introduced two key enhancements, namely imitation and delayed adversarial training, to further enhance robustness. Through extensive experiments on popular datasets like PASCAL-VOC and MS-COCO, we have demonstrated the efficacy of our method in achieving robust object detection.


{\small
\bibliographystyle{abbrvnat}
\setcitestyle{authoryear,open={((},close={))}} 

\bibliography{refs}
}

\section{Extended Results}

Here, we provide additional insights into our method by presenting extended qualitative results and discussing common errors that our method can encounter. Figure~\ref{tab:errors} illustrates a comprehensive list of these common mistakes, which include instances where the model misses some objects, assigns incorrect labels to correct bounding boxes, generates slightly larger bounding boxes, fails to detect small objects, or does not detect any objects at all. Furthermore, in Figure~\ref{tab:comparison}, we showcase our extended results to further support our findings and analysis.

\begin{table}[ht!]
    \centering
    \begin{tabular}{cc}
       \includegraphics[width=0.5\textwidth]{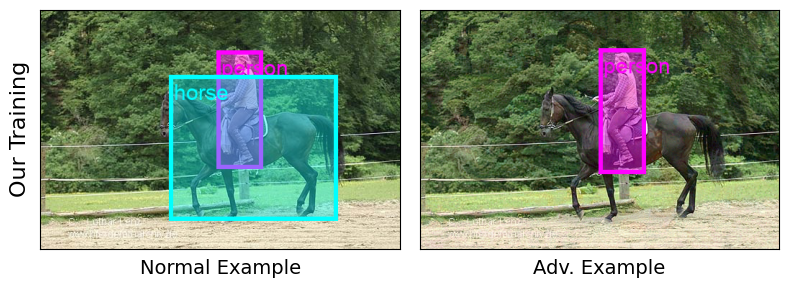}  &        \includegraphics[width=0.3\textwidth]{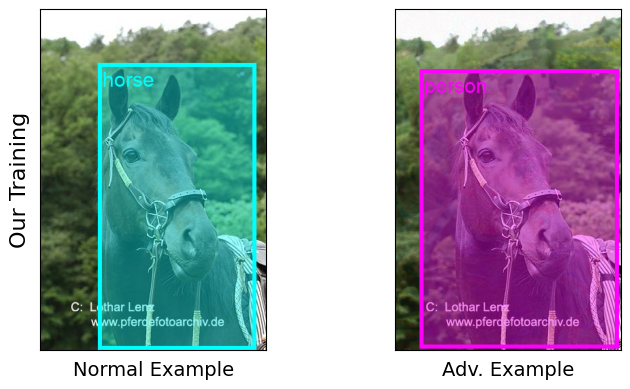} \\
         \includegraphics[width=0.5\textwidth]{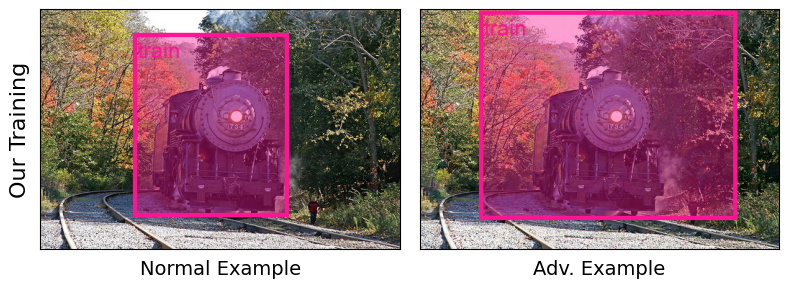} &
         \includegraphics[width=0.45\textwidth]{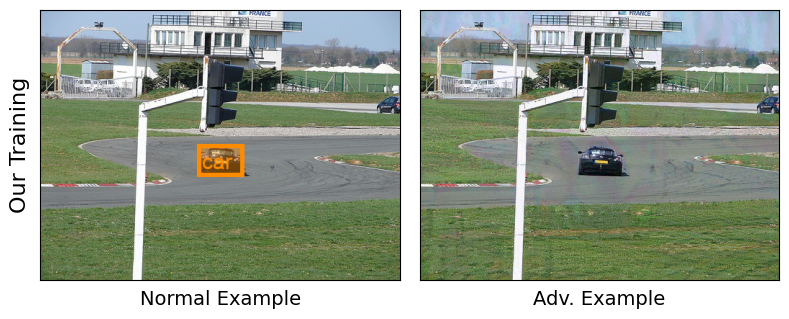} \\
      \includegraphics[width=0.5\textwidth]{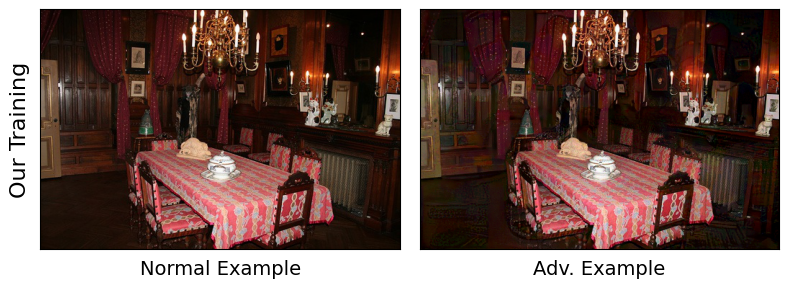} &
          \includegraphics[width=0.5\textwidth]{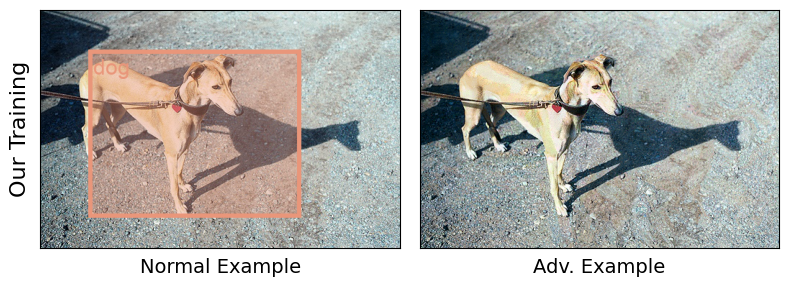}
    \end{tabular}
    \caption{Common errors made by a model trained via. our proposed method.}
    \label{tab:errors}
\end{table}

\begin{table*}[]
    \centering
    \begin{tabular}{cc}
       \includegraphics[width=0.5\textwidth]{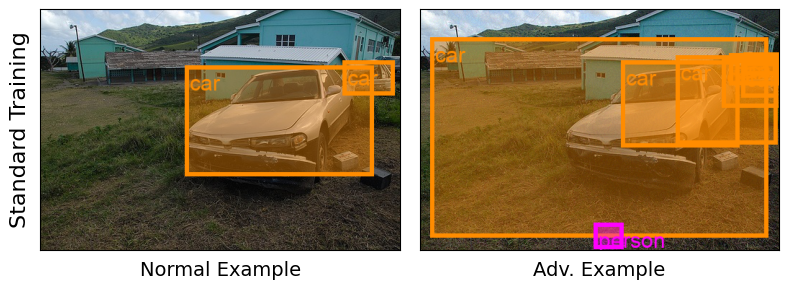} & \includegraphics[width=0.5\textwidth]{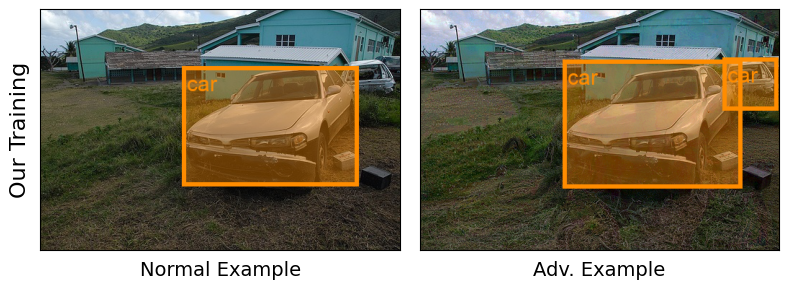}   \\
    \includegraphics[width=0.5\textwidth]{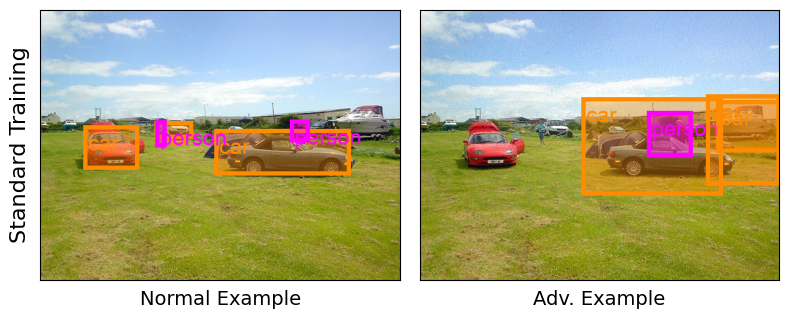} & \includegraphics[width=0.5\textwidth]{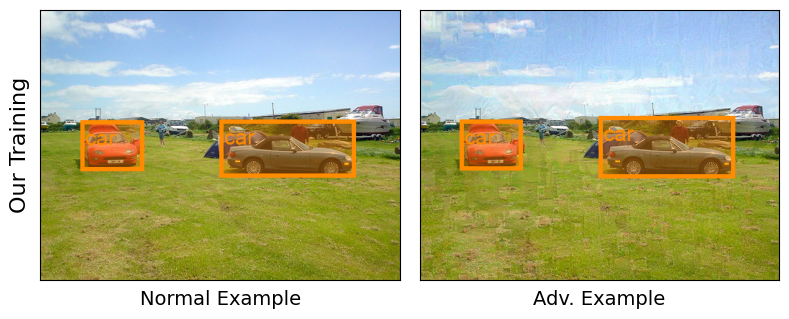}   \\
        \includegraphics[width=0.5\textwidth]{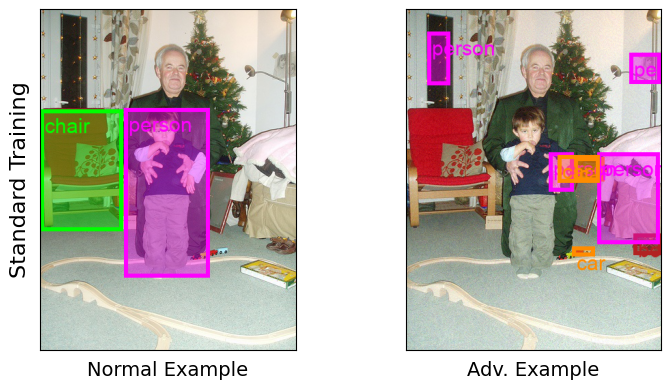} & \includegraphics[width=0.5\textwidth]{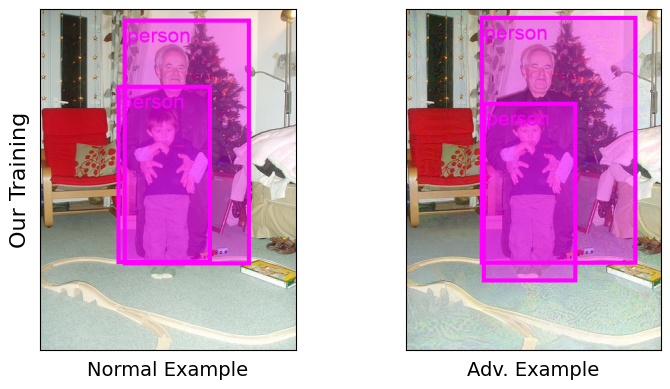}   \\
        \includegraphics[width=0.5\textwidth]{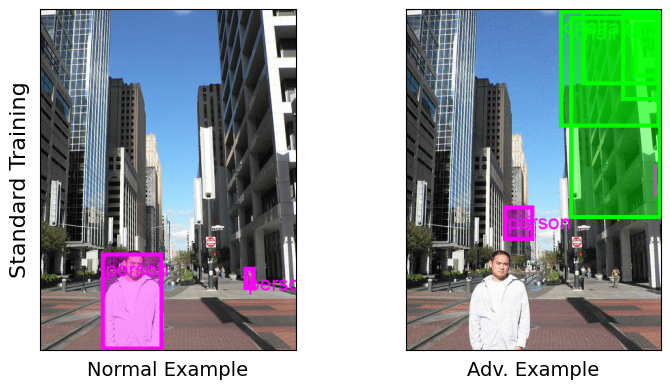} & \includegraphics[width=0.5\textwidth]{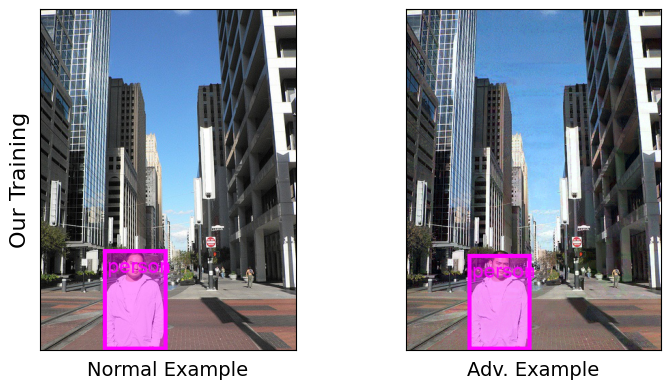}   \\
        \includegraphics[width=0.5\textwidth]{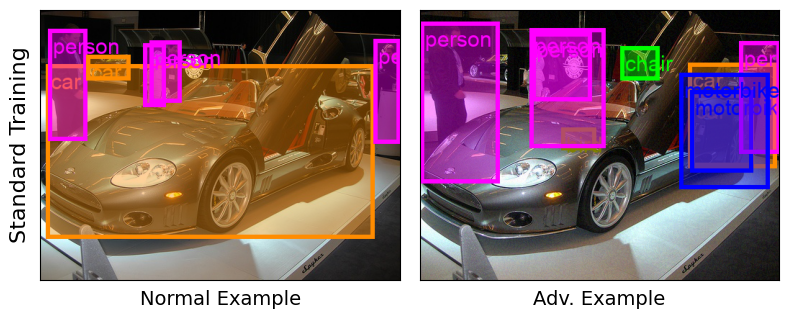} & \includegraphics[width=0.5\textwidth]{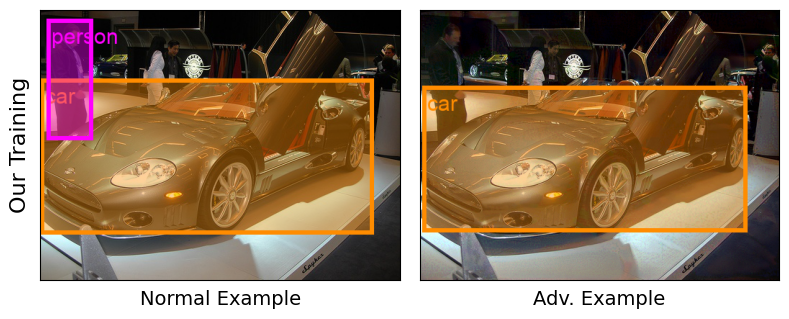}   \\
       \includegraphics[width=0.5\textwidth]{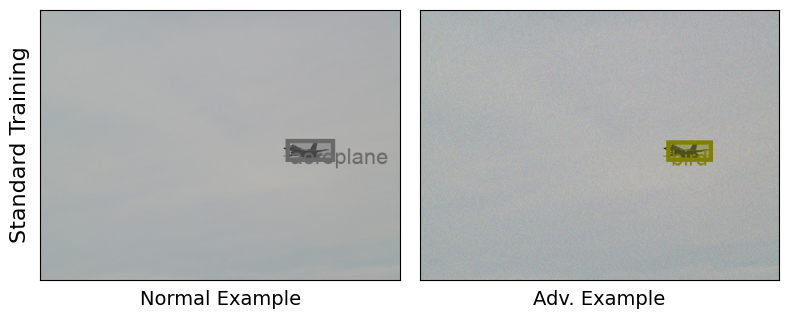} & 
       \includegraphics[width=0.5\textwidth]{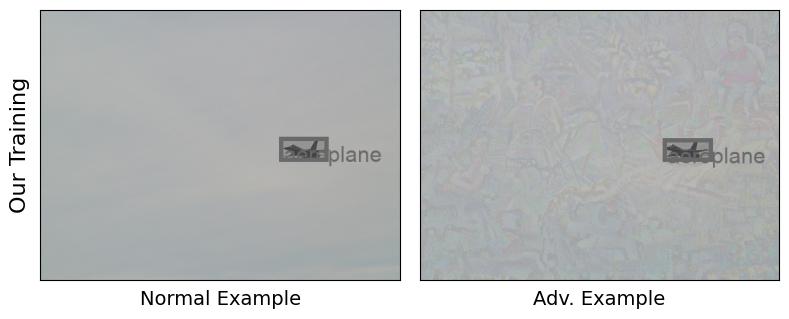}   \\
    \end{tabular}
    \caption{A comparison of our method vs. standard training for normal and adversarial examples. }
    \label{tab:comparison}
\end{table*}

\end{document}